\documentclass{article}
\usepackage{ijcai23}

\usepackage{times}
\usepackage{soul}
\usepackage{url}
\usepackage[hidelinks]{hyperref}
\usepackage[utf8]{inputenc}
\usepackage[small]{caption}
\usepackage{graphicx}
\usepackage{amsmath}
\usepackage{amsthm}
\usepackage{booktabs}
\usepackage[switch]{lineno}

\usepackage{paralist}
\usepackage{booktabs}
\usepackage{amsfonts}
\usepackage{CJKutf8}
\usepackage{subfigure} 
\usepackage[ruled,vlined]{algorithm2e}
\usepackage{amssymb}
\usepackage{enumitem}
\usepackage{setspace}%行间距
\usepackage{amsmath} 
\usepackage{amssymb}
\usepackage{xcolor}
\usepackage{times}
\usepackage{latexsym}
\usepackage{multirow}
\usepackage{multicol}

\title{Less Learn Shortcut: Analyzing and Mitigating Learning of \\ Spurious Feature-Label Correlation}

\author{
Yanrui Du$^1$ \thanks{\llap{}\:\:\:The work was done when Yanrui Du was doing internship at Baidu.}
\and
Jing Yan$^2$
\and
Yan Chen$^2$
\and
Jing Liu$^2$
\and
Sendong Zhao$^1$\thanks{\llap{}\:\:\:Corresponding author}
\and \\
Qiaoqiao She$^2$
\and
Hua Wu$^2$
\and
Haifeng Wang$^2$
\and
Bing Qin$^1$
\affiliations
$^1$Harbin Institute of Technology, Harbin, China\\
$^2$Baidu Inc., Beijing, China\\
\emails
\{ yrdu, sdzhao, bqin\}@ir.hit.edu.cn\\
\{yanjing09, chenyan22, liujing46, sheqiaoqiao, wu\_hua, wanghaifeng\}@baidu.com
}

\begin{document}

\maketitle
\begin{abstract}
Recent research has revealed that deep neural networks often take dataset biases as a shortcut to make decisions rather than understand tasks, leading to failures in real-world applications. In this study, we focus on the spurious correlation between word features and labels that models learn from the biased data distribution of training data. In particular, we define the word highly co-occurring with a specific label as biased word, and the example containing biased word as biased example. Our analysis shows that biased examples are easier for models to learn, while at the time of prediction, biased words make a significantly higher contribution to the models' predictions, and models tend to assign predicted labels over-relying on the spurious correlation between words and labels. To mitigate models' over-reliance on the shortcut (i.e. spurious correlation), we propose a training strategy Less-Learn-Shortcut (LLS): our strategy quantifies the biased degree of the biased examples and down-weights them accordingly. Experimental results on Question Matching, Natural Language Inference and Sentiment Analysis tasks show that LLS is a task-agnostic strategy and can improve the model performance on adversarial data while maintaining good performance on in-domain data.

\end{abstract}

\begin{CJK*}{UTF8}{gbsn}
\section{Introduction}

%大背景 %列举一个具体的例子
Pre-trained language models, e.g. BERT~\cite{devlin2018bert}, ERNIE~\cite{sun2019ernie} and RoBERTa~\cite{liu2019roberta}, have achieved great success on many NLP tasks. 
However, recent studies highlighted that pre-trained models tend to take dataset biases as a shortcut, rather than truly understand tasks~\cite{schuster2019towards,niven2019probing}. Models' over-reliance on the shortcut results in their poor generalization ability and low robustness~\cite{geirhos2020shortcut}.

The phenomenon of shortcut learning has been widely studied in various NLP tasks. Many previous studies examine this phenomenon by constructing artificial adversarial examples, and employ adversarial data augmentation to enhance model robustness~\cite{jia2017adversarial,alzantot2018generating,ren2019generating,jin2020bert}. These studies reported high success rates on artificial adversarial examples, but it is uncertain if the models will perform well on real-world data distributions~\cite{morris2020reevaluating,bender2020climbing}. Additionally, recent work~\cite{balkir2022challenges} indicated that few studies have applied explainable methods to understand or investigate the impact of shortcut learning.

% dataset biases.
Previous works point out that shortcuts can be traced back to dataset biases~\cite{lai2021machine,gururangan2018annotation,kavumba2021learning,Du2021TowardsIA,mccoy2019right,liu2019inoculation,kavumba2021learning,goyal2017making,ye2021case,dawkins2021marked}. For example, if ``not" happens to be \emph{contradiction} for most of the training data in Natural Language Inference (NLI) tasks, detecting ``not" becomes a successful strategy for models' prediction, thus leading to an unexpected performance on a shift distribution~\cite{gururangan2018annotation}. However, most studies are limited to analyzing task-specific shortcuts, which are prohibitive to be transferred to other tasks.

In this work, we analyze the correlations between simple features (e.g. words) and labels, which can be originated from the biased data distribution of any NLP task, to quantitatively investigate the shortcut learning behavior of NLP models. Existing work has argued that, for any NLP task, no single feature on its own should contain information about the labels, and any correlation between simple features and labels is spurious~\cite{gardner2021competency}. Based on the above analysis, we propose a task-agnostic training strategy \textit{Less-Learn-Shortcut (LLS)}, which mitigates the shortcut behavior of models, thereby improving their performance on adversarial data.

To examine the spurious feature-label correlation, we first introduce two definitions: biased word, which is the word highly co-occurring with a specific label in a dataset, and biased example, which is the example containing at least one biased word. Then we quantitatively analyze the spurious feature-label correlations on the Question Matching (QM) task. Based on our analysis, we propose our training strategy \textit{LLS}, with which biased training examples are down-weighted according to their biased degrees, and the models' over-reliance on the biased words is penalized during fine-tuning. We conduct extensive experiments on QM, NLI and Sentiment Analysis (SA) tasks to evaluate our training strategy and compare it to other task-agnostic strategies such as \textit{Rew.$_{bias}$} ~\cite{utama2020towards,clark2019don} and \textit{Forg.}~\cite{yaghoobzadeh2021increasing}. Our experimental results demonstrate that \textit{LLS} can improve the model performance on adversarial data while maintaining good performance on in-domain data, and can be easily transferred to different NLP tasks. Additionally, we explore the scenarios in which the above strategies are applicable.

In general, we have the following major findings and contributions:
\begin{itemize}[leftmargin=*,noitemsep,topsep=0pt]
    \item We reveal that biased examples (as defined in Sec.~\ref{definition}) are easier to be learned than other examples, and with an explainable method LIME~\cite{ribeiro2016should}, we find that biased words make significantly higher contributions to models' predictions than random words (see Sec.~\ref{bias_training_loss}). 
    \item We find that biased words will affect models' predictions, and that models tend to assign labels highly correlated to the biased words (see Sec.~\ref{how_bias_affect}). 
    \item To mitigate the models' over-reliance on the spurious correlation, we propose a training strategy \textit{Less-Learn-Shortcut (LLS)}. Experimental results show that \textit{LLS} can improve the models' performance on adversarial data while maintaining good performance on in-domain data. Furthermore, we compare \textit{LLS} to existing strategies and reveal their respective applicable scenarios. (see Sec.~\ref{how_to_migiate}).
\end{itemize}

\section{Preliminary}\label{set_up}

In this section, we first introduce the QM datasets on which we analyze the spurious feature-label correlation, then we give the definitions of \textit{biased word} and \textit{biased example}. At last, we provide the settings of our experiments.

\subsection{Datasets}
We conduct our analysis on three datasets, LCQMC, DuQM and OPPO\footnote{The datasets can be downloaded on https://luge.ai.}, all of which are about QM task and collected from real-world applications. LCQMC~\cite{liu2018lcqmc} is a large-scale Chinese question matching corpus proposed by Harbin Institute of Technology in the general domain BaiduZhidao. 
DuQM~\cite{zhu2021duqm}
%~\footnote{https://github.com/baidu/DuReader/tree/master/DuQM.} 
is a fine-grained controlled adversarial dataset aimed to evaluate the robustness of QM models and generated based on the queries collected from Baidu Search Engine~\footnote{http://www.baidu.com.}. OPPO is collected from OPPO XiaoBu Dialogue application and can be downloaded on CCF Big Data \& Computing Intelligence Contest. The data statistics are provided in Tab.~\ref{tab:dataset_basic_statics} (in App.~\ref{app:lcqmc statics}).

\subsection{Definitions}\label{definition}
Here we provide the definitions we will use in this work.
If we denote $W$ as all words in the dataset, the set of examples containing a specific word $w_i$ can be formalized as $S(w_i)$, and the frequency of $w_i$ can be formalized as $f_{w_{i}}$. We define \emph{biased degree} as $d_{w_i}^{c_m}$  to measure the degree of word $w_i$ co-occurring with category $c_m$ (for QM task, $c_m \in (0,1)$) and it can be denoted as
\begin{equation}
d_{w_i}^{c_m} = \frac{|S(w_i, c_m)|}{|S(w_i)|} =  \frac{|S(w_i, c_m)|}{f_{w_{i}}}
\end{equation}
where $|S(w_i, c_m)|$ represents the number of examples with $w_i$ and labeled with $c_m$.

\begin{table}[t]
\centering
\begin{spacing}{1.2}
% \scalebox{0.55}
% \small
{
\begin{tabular}{l|cccc}
\toprule[0.7pt]
{\textbf{\# Word}} & {\textbf{\# B-word$_{0}$}} & {\textbf{\# B-word$_{1}$}} & {\textbf{\# B-word}} \\
\midrule[0.5pt]
% & {27.24\%}
 {58,230} & {11,145} & {4,721} & {15,866} \\
\bottomrule[0.7pt]
\end{tabular}
}
\caption{The statistics of biased words in LCQMC$_{train}$.}
\label{tab:bias_word_statics}
\end{spacing}
\end{table}

\begin{table}[t]
\begin{spacing}{1.1}
\centering
\small
{
\begin{tabular}{l|c|c|c}
\toprule[0.7pt]
{\textbf{Dataset}} & {\textbf{\#\ Examples}} & {\textbf{\#\ B-exp}} & {\textbf{\%B-exp}}\\
\midrule[0.7pt]
{LCQMC$_{train}$} & {238,766} & {98,260} & {41.15\%} \\
\midrule[0.5pt]
{LCQMC$_{test}$} & {12,500} & {3,246} & {25.97\%} \\
{DuQM} & {10,121} & {3,264} & {32.25\%} \\
{OPPO} & {10,000} & {2,498} & {24.98\%} \\
\bottomrule[0.7pt]
\end{tabular}
}
\caption{The statistics of biased examples in the datasets. B-exp represents biased example.}
\label{tab:bias_example_statics}
\end{spacing}
\end{table}
\paragraph{Biased word.} A word highly correlated with a specific label in a dataset.\footnote{Word is the smallest independent lexical item with its own objective or practical meaning. We use Lexical Analysis of Chinese~\cite{jiao2018chinese} (https://github.com/baidu/lac) for word segmentation in this work.} To better discuss it, we define \textit{biased word} as the word $w_i$ with $f_{w_{i}}\geq3$ and $d_{w_i}^{c_m}\geq0.8$ for QM task in Sec. 2 and 3. It is worth mentioning that the biased words we analyze in this work are originated from the training set.

We further define \emph{biased word$_0$} and \emph{biased word$_1$} as the words highly correlated to category $0$ and $1$. As shown in Tab.~\ref{tab:bias_word_illustraion} (in App.~\ref{app:example of bias-word}), ``简便" (``handy") occurs in 35 examples, 33 of which are with category 1, hence it is a biased word$_1$. Tab.~\ref{tab:bias_word_statics} shows that 27.24\% ($15864/58230$) of words are biased words, and there are more biased word$_0$ than biased word$_1$ in LCQMC$_{train}$.

\paragraph{Biased example.} An example containing at least one biased word. As shown in Tab.~\ref{tab:bias_example_statics}, 41.15\% of examples in LCQMC$_{train}$ are biased examples, which are 25.97\%, 32.25\% and 24.98\% in LCQMC$_{test}$, DuQM and OPPO respectively. Since the biased words occur in almost half of the examples in LCQMC$_{train}$, it is meaningful to study their effects on models. The examples without biased words are defined as \textit{unbiased example}.

\begin{figure}[ht]
\centering
\subfigure[Training loss curve of RoBERTa.]{
    \centering
    \includegraphics[scale=0.32]{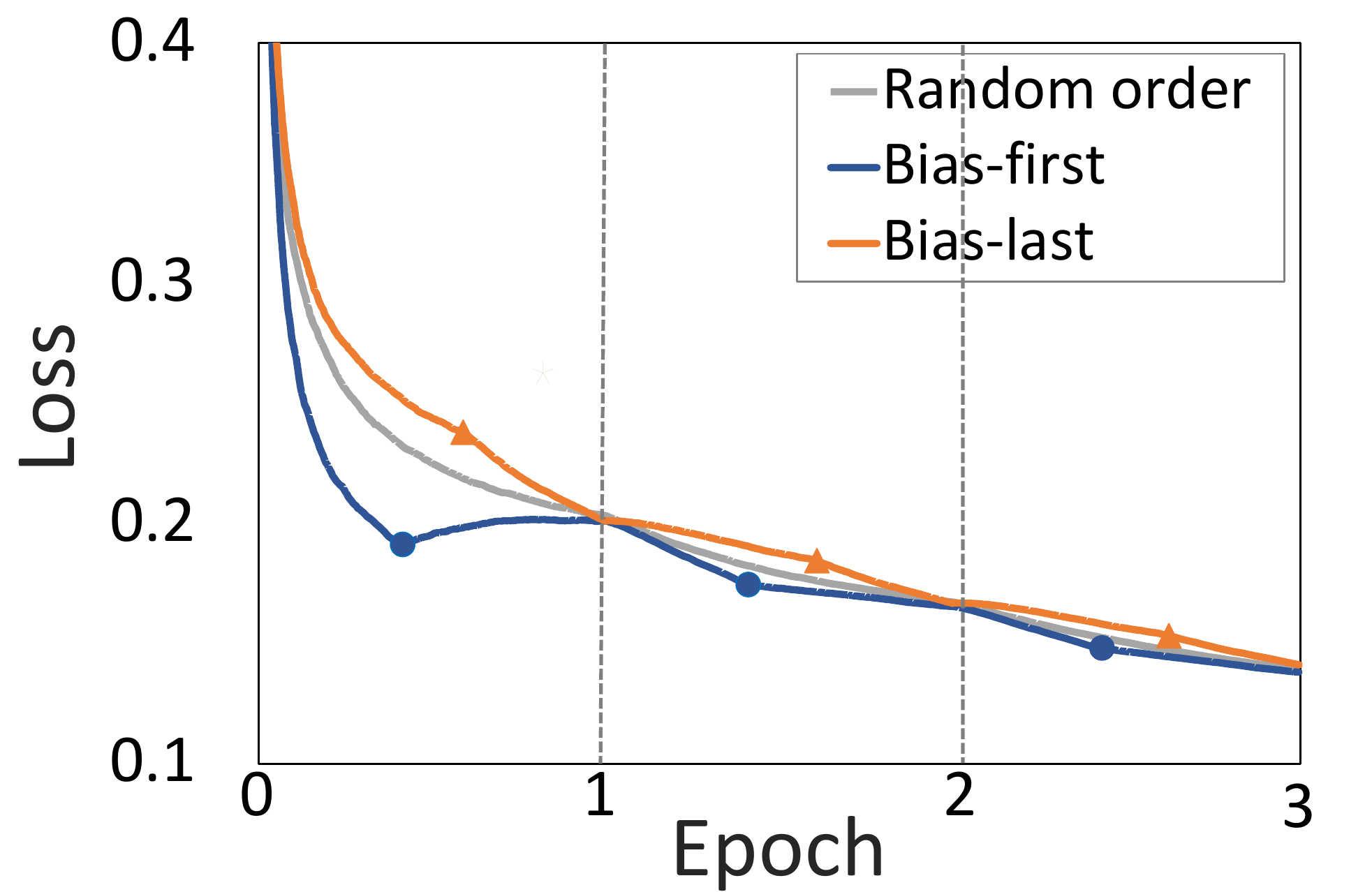}
}
\caption{Training loss curves of RoBERTa on LCQMC$_{train}$, in which \textcolor{blue}{$\bullet$} represents the time of finishing learning biased examples, and \textcolor{orange}{$\blacktriangle$} represents the time of finishing learning unbiased examples.}
\label{fig:training loss curve}
\end{figure}

\subsection{Experimental Setup}\label{experimental_setup}
\paragraph{Models.}
We conduct our experiments on three popular publicly available pre-trained models,
BERT-base\footnote{https://github.com/google-research/bert.}, ERNIE$_{1.0}$\footnote{https://github.com/PaddlePaddle/ERNIE.} and RoBERTa-large\footnote{https://github.com/ymcui/Chinese-BERT-wwm.}.
\paragraph{Metrics.}As most of the classification tasks, we use accuracy to evaluate the performance of models.
\paragraph{Training details.} We use the integrated interface BertForSequenceClassification\footnote{https://huggingface.co/docs/transformers/.} from huggingface for our experiment and use different learning rates for different pre-trained models. Specifically, for RoBERTa$_{large}$, the learning rate is 5e-6. For BERT$_{base}$ and ERNIE$_{1.0}$, the learning rate is 2e-5. The proportion of weight decay is 0.01 and the batch size is 64. We train two epochs for BERT$_{base}$ and ERNIE$_{1.0}$, and train three epochs for RoBERTa$_{large}$. Every 500 steps, we check the performance of models on LCQMC$_{dev}$ and choose the checkpoint with the highest accuracy as our main model, and report average results with three different seeds on LCQMC$_{test}$, DuQM and OPPO.

% \section{Effect of feature-label correlation on the model}\label{chapter: influenece of bias}
% training loss

\section{Effect of Feature-Label Correlation}\label{chapter: influenece of bias}

% %关注度
\begin{figure}[ht]
\centering
\subfigure[Results on LCQMC$_{test}$.]{
\centering
\includegraphics[scale=0.24]{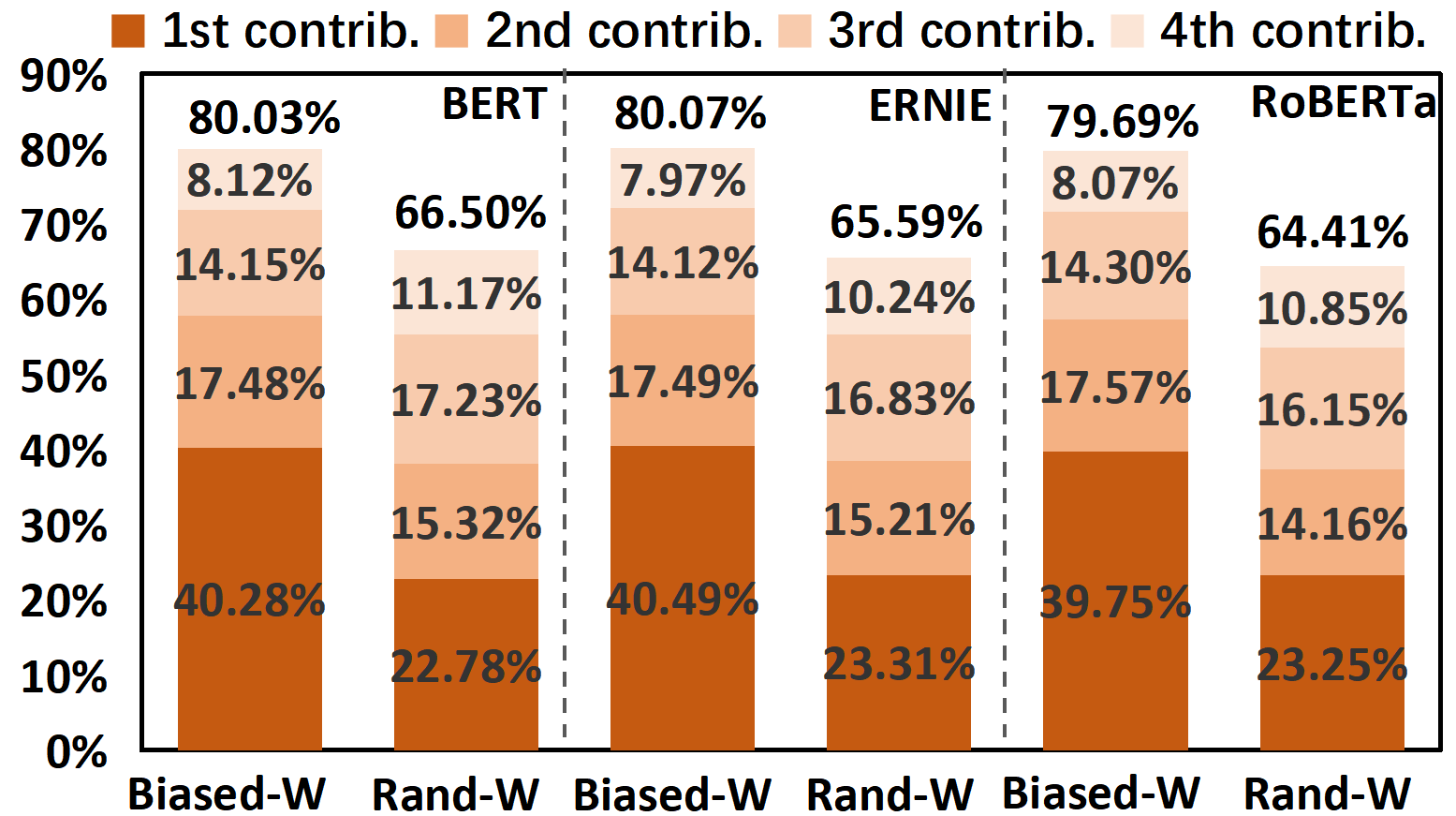}
}

\centering
\subfigure[Results on DuQM.]{
\centering
\includegraphics[scale=0.24]{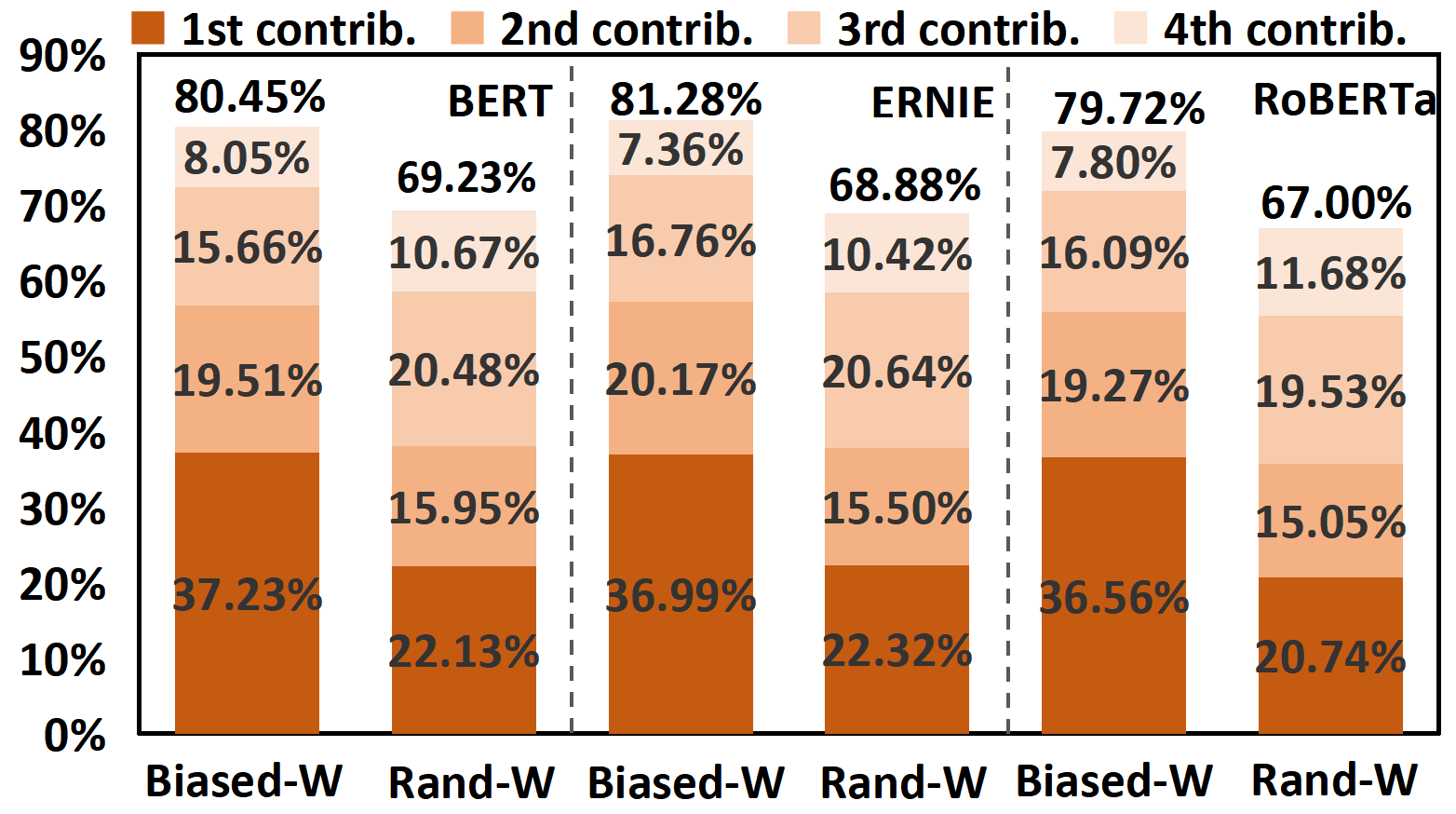}
}

\centering
\subfigure[Results on OPPO.]{
\centering
\includegraphics[scale=0.24]{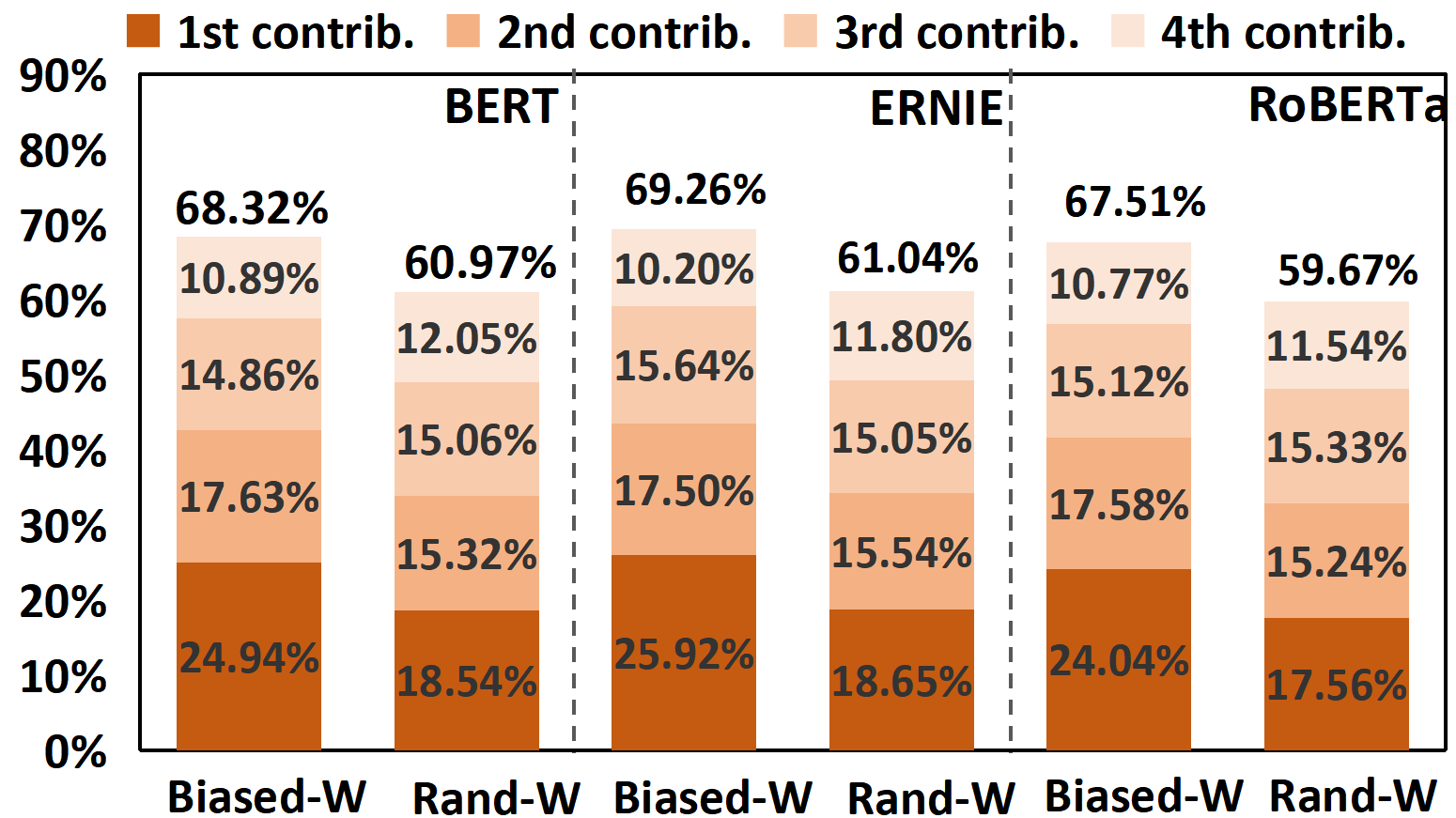}
}
\caption{Probability of biased words and random words with the 1st, 2nd, 3rd, 4th contribution on three test sets. Bias-W and Rand-W represent biased words and random words respectively.}
\label{fig:attention figure}
\end{figure}

% Bias-word represents biased word.

%预测趋势
\begin{figure*}[ht]
\centering
\subfigure[Biased word$_0$ on LCQCM$_{test}$.]{
\includegraphics[scale=0.28]{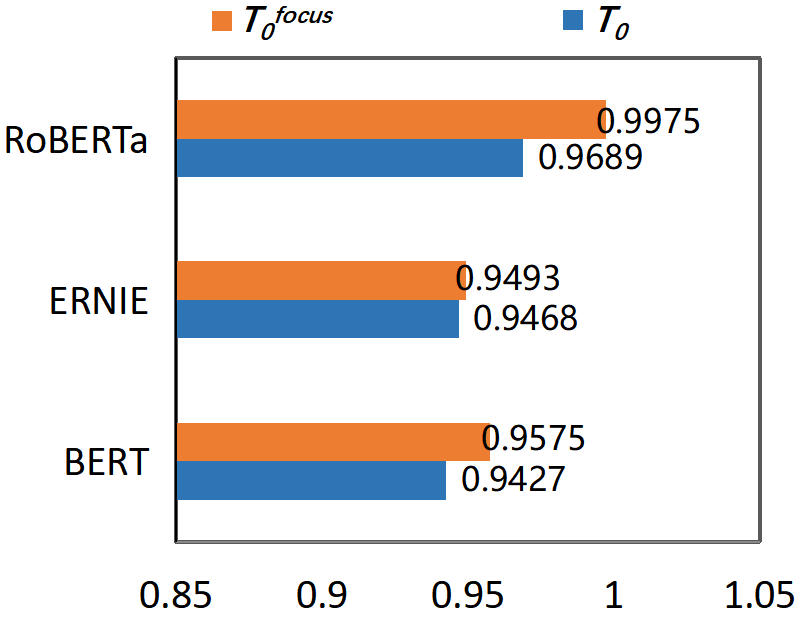}
\label{fig:lcqmc_0}

}
\subfigure[Biased word$_0$ on DuQM.]{
\includegraphics[scale=0.28]{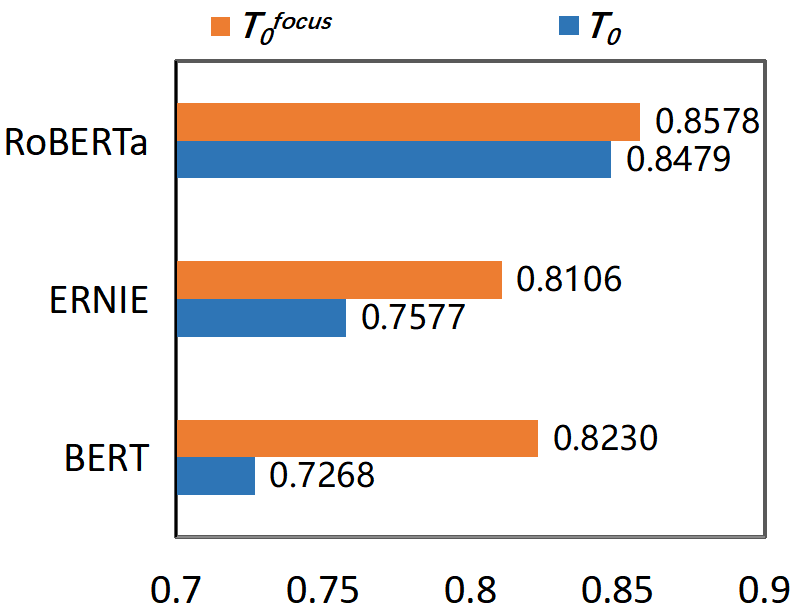}
\label{fig:duqm_0}
}
\subfigure[Biased word$_0$ on OPPO.]{
\includegraphics[scale=0.28]{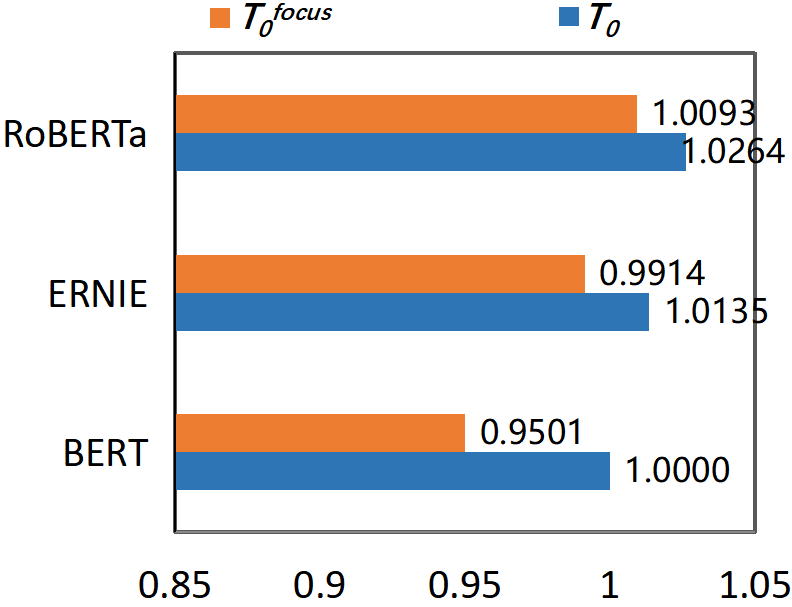}
\label{fig:oppo_0}
}
\subfigure[Biased word$_1$ on LCQCM$_{test}$.]{
\includegraphics[scale=0.28]{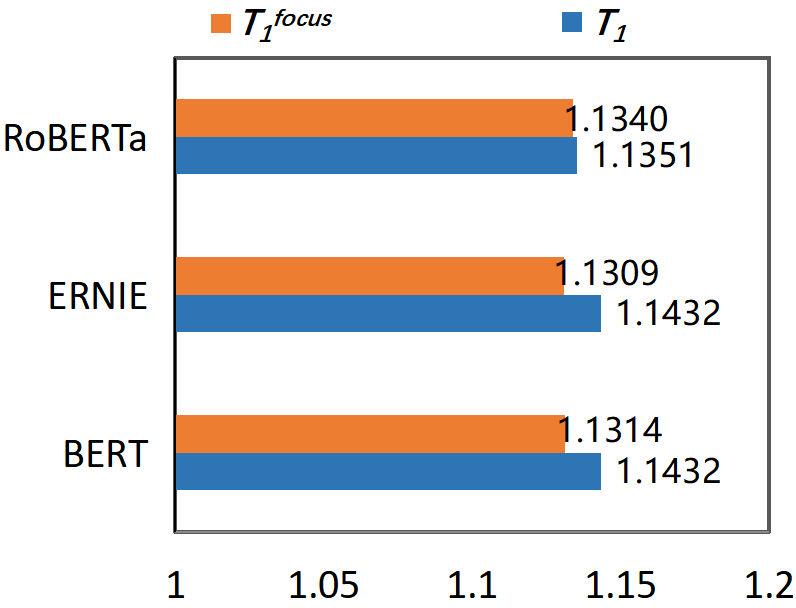}
\label{fig:lcqmc_1}
}
\subfigure[Biased word$_1$ on DuQM.]{
\includegraphics[scale=0.28]{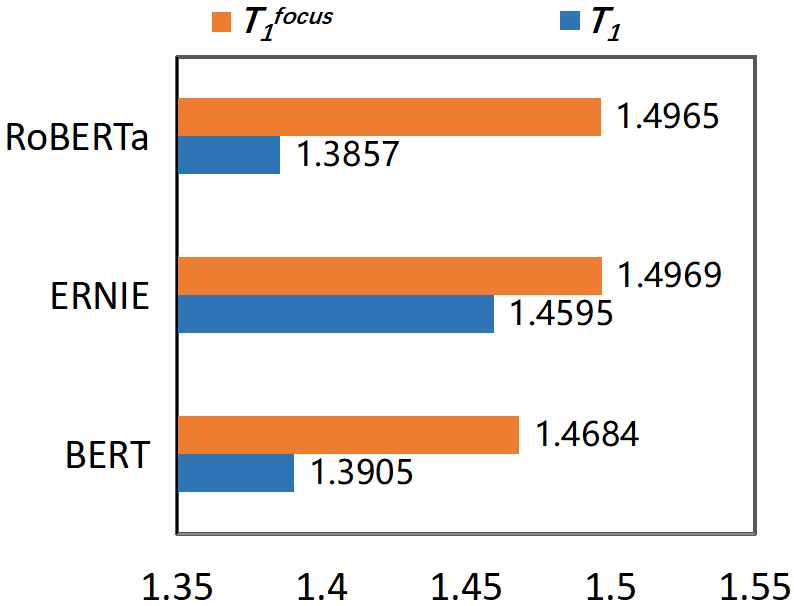}
\label{fig:duqm_1}
}
\subfigure[Biased word$_1$ on OPPO.]{
\includegraphics[scale=0.28]{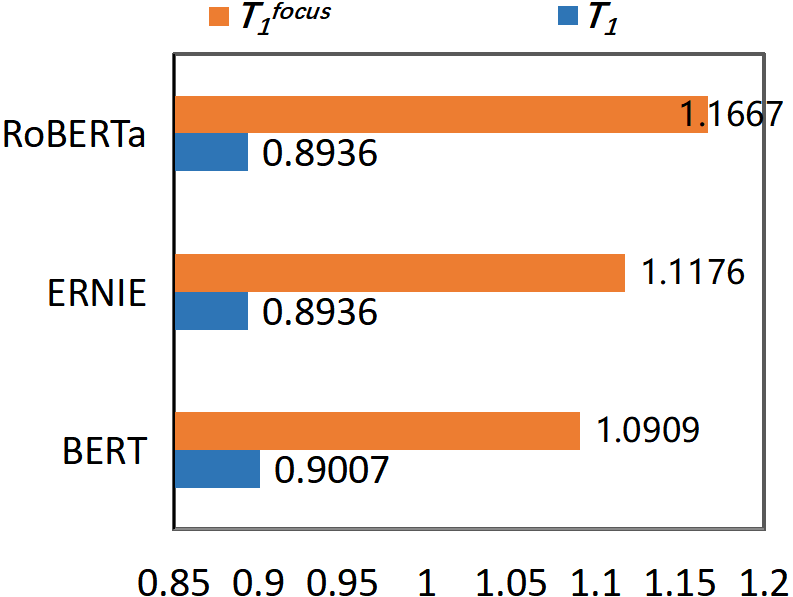}
\label{fig:oppo_1}
}
\caption{Tendency to predict 0 of biased word$_0$ and predict 1 of biased word$_1$.}
\label{fig:influence_of_bias}
\end{figure*}

The dataset statistics in Sec.~\ref{set_up} show that 41.15\% of examples in LCQMC$_{train}$ contain biased words. It is a reasonable assumption that the spurious feature-label correlations would affect the models' behavior and performance. To validate our assumption: 1) we conduct a behavior analysis of the model's learning 
%(See Sec.~\ref{bias_training_loss})
and deciding (See Sec.~\ref{bias_training_loss}); 2) we discuss how the feature-label correlation affects the models’ performance by probing the relationship between the biased word and the predicted label (See Sec.~\ref{how_bias_affect}). In Sec.~\ref{sec:word-overlap bias} we discuss another type of shortcut \emph{word-overlap} and argue that different shortcuts may interact together. 
% training loss
\subsection{Feature-Label Correlation and Models' Behavior}\label{bias_training_loss}

\paragraph{Models' learning.}
To observe the models' behavior during training, we separate LCQMC$_{train}$ into two subsets, biased examples and unbiased examples, and reorganize the train examples in 3 orders:
\begin{itemize}[leftmargin=*,noitemsep,topsep=0pt]
\item \textit{bias-first}: firstly biased examples, then unbiased examples;
\item \textit{bias-last}: firstly unbiased examples, then biased examples;
\item \textit{random order}: shuffle the examples randomly.
\end{itemize}

We finetune three models (BERT, ERNIE and RoBERTa) in above three orders and plot the training loss curves in Fig.~\ref{fig:training loss curve} and Fig.\ref{fig:training loss curve app} (see in App ~\ref{app:training loss}). The training loss curves of all three models present the same tendencies:
\begin{itemize}[leftmargin=*,noitemsep,topsep=0pt]
\item If bias-first, for each epoch, the loss curve drops more rapidly than random order. After learning all the biased examples, the loss curve rises slightly and then decreases.
\item If bias-last, the tendency is contrary: for each epoch, the loss drops more slowly than random order until all the unbiased examples have been learned, and then the curve decreases faster.
\end{itemize}

The above observations reflect that models behave differently when they learn biased examples and unbiased examples: the loss curves of biased examples drop more sharply than other examples, which indicates that the words highly correlated with specific labels are relatively easier for models to learn, and the correlations between words and labels are captured by models as shortcuts. 
%Generally, we validate our assumption that the feature-label correlations affect the model's learning. 

%\subsection{Shortcuts and models' prediction}\label{bias_and_predict}
\paragraph{Models' deciding.}\label{bias_and_predict}
In this part, we provide a quantitative analysis of the spurious feature-label correlation's impact on models' deciding. If it is easier for a model to learn, will the biased words make greater contributions when predicting?
%As introduced in Section~\ref{set_up}, 
Here we select LIME method to measure the contributions of different words in one input to the final prediction, which can interpret the models' prediction based on locally approximating the model around a given prediction. 

To observe the contributions of the biased words, we rank the words based on their contribution scores computing with LIME method. In Fig.~\ref{fig:attention figure}, we illustrate the ratios of biased words with the highest, second, third, and fourth contribution in three test sets. For comparison, we randomly select words from the input excluding stop-words as random word baseline, and also plot their ratios in Fig.~\ref{fig:attention figure}. Compared to the random words, the biased words have significantly higher ratios to be ranked among the highest 4, which is about 80\% in LCQMC$_{test}$ and DuQM, 68\% in OPPO.

% As we reported in Tab.~\ref{tab:dataset_basic_statics}, in these three test sets, each sample is composed of around 9 to 11 words. 

% To observe the contribution of Biased word in each test sample, we rank the words based on their contributions computing with LIME method. In Fig.~\ref{fig:attention figure}, we illustrate the probabilities of Biased words with the highest, second, third, and fourth contribution in three test sets. For comparison, we select some words randomly as baseline and observe the contribution of random-word. Moreover, as we know, stopwords tend not to have a large impact on model's predictions. So, we select random-words except for stopwords as enhanced baseline and plot above probabilities in Fig.~\ref{fig:attention figure}.
% \textcolor{red}{增加:去除停用词，看关注度，重新画图}

%\paragraph{Results analysis.} Fig.~\ref{fig:attention figure} shows that the random words have a probability of around 66\% ranked among the highest 4 contribution words. Compared with random words, the biased words have significantly higher probability to be ranked among the highest 4, which is about 80\% in LCQMC$_{test}$ and DuQM, 68\% in OPPO. Specifically, in about 40\% of biased examples of LCQMC$_{test}$, 37\% of biased examples of DuQM, and 25\% of biased examples of OPPO, the Biased word is the word with the highest contribution to final prediction, which is 1.5\textasciitilde2 times to random words (which is only 17\%\textasciitilde23\%). 

In summary, the biased examples are easier for models to learn, and the biased words make significantly higher contributions than random words, which implies that models tend to pay more attention to biased words when deciding. With the analysis in this section, we can conclude that the biased word is a shortcut for the models and will affect the models' behavior. It is therefore substantial to further analyze how it affects the models. %Therefore, it sparks our interest in studying how bias affects the models.

\begin{table*}[ht]
\begin{spacing}{1.2}
\centering
% \scalebox{0.5}
\small
{
\begin{tabular}{lcc|cc|cc|cc|cc}
\toprule[0.7pt]
\multirow{2}{*}{\textbf{Model}} & \multicolumn{2}{c|}{\textbf{Dist. $ \leqslant 1$}} & \multicolumn{2}{c|}{\textbf{Dist. $ \leqslant 2$}} & \multicolumn{2}{c|}{\textbf{Dist. $ \leqslant 3$}} & \multicolumn{2}{c|}{\textbf{Dist. $ \leqslant 4$}} & \multicolumn{2}{c}{\textbf{Dist. $ \leqslant 5$}}\\

{} & $T_{0}$ & $T_{0}^{focus}$  & $T_{0}$ & $T_{0}^{focus}$  & $T_{0}$ & $T_{0}^{focus}$  & $T_{0}$ & $T_{0}^{focus}$  & $T_{0}$ & $T_{0}^{focus}$\\

\midrule[0.7pt]
BERT & 0.739 & 0.833 & 0.765 & 0.821 & 0.800 & 0.847 & 0.866 & 0.909  & 0.910 & 0.933 \\
ERNIE & 0.761 & 0.857 & 0.779 & 0.786 & 0.841 & 0.847 & 0.894 & 0.905 & 0.946 & 0.970\\
RoBERTa & 0.870 & 0.938 & 0.875 & 0.932 & 0.905 & 0.935 & 0.950 & 0.978 & 0.984 & 1.004 \\

\midrule[0.7pt]
 \multicolumn{1}{c}{$\overline{\Delta}$} & \multicolumn{2}{c|}{0.086} & \multicolumn{2}{c|}{0.040} &
\multicolumn{2}{c|}{0.028} &
\multicolumn{2}{c|}{0.027} &
\multicolumn{2}{c}{0.022} \\
\bottomrule[0.7pt]
\end{tabular}
}
\caption{Tendency to predict 0 with edit distance less than 6. $\overline\Delta$ denotes the mean of $T_{0}^{focus}$-$T_0$ on BERT, ERNIE and RoBERTa.}

\label{tab:oppo_result}
\end{spacing}
\end{table*}
%回答第二个问题
%有偏性确实会影响模型的决策，但会如何影响呢？我们猜测是偏0词会使得决策偏0，偏1词会使得决策偏1.
%这章要解决的问题是有偏词是如何影响决策的，从有偏词对模型的表现的影响出发
\subsection{Feature-Label Correlation and Models’ Prediction}\label{how_bias_affect}
% Previous works show that superficial cues exist in many data sets and are widely studied~\cite{bolukbasi2016man,may2019measuring,ravfogel2020null,webster2020measuring,kaneko2021debiasing}. However, there are few quantitative analysis to discuss how these cues affect the model's decision. We have proved that Biased words tend to make more contributions to the final prediction than other words. In this section, we will focus on the examples where Biased words make the \textbf{greatest} contribution to the final prediction, in which the effect of Biased word would be more significant, to probe the relationship between the Biased word and the predicted category. A reasonable guess is that the models tend to assign the category highly relying on the distribution bias trick, i.e. a Biased word$_1$ with high contribution to the final decision will bring a prediction of category $1$ and vice versa. 

Existing works focus only on verifying the existence of shortcut~\cite{bolukbasi2016man,may2019measuring,ravfogel2020null,webster2020measuring,kaneko2021debiasing}. However, there are few quantitative analyses to discuss how the shortcut affects the models' predictions exactly. In this part, we will focus on probing the relationship between the biased word and predicted label to analyze how the spurious feature-label correlations impact models. As the biased words are highly correlated to a specific label, it is a reasonable guess that the models tend to assign predicted labels highly correlated to the biased words. 

% A reasonable guess is that the models tend to assign the category highly relying on the distribution bias trick, i.e. a Biased word$_1$ with high contribution to the final decision will bring a prediction of category $1$ and vice versa.  

% For two-classification task, when $c_m$ is 1 (positive label), $T_{c_m}$ represents true positive rate; when $c_m$ is 0 (negative label), $T_{c_m}$ represents true negative rate. $T_{c_m}$ can also be applied in multi-classification scenarios.
% \qquad c_m \in (0,1)
% $c_m \in (0,1)$ for two-classification task,
% by 5\%\textasciitilde7\%

%介绍实验：分析偏0字/词的实验
Although the biased words tend to contribute more (discussed in Sec.~\ref{bias_and_predict}), not all biased words make great contribution during predicting. To better analyze the impact of biased word on predicted label, we focus on the testing examples where biased word contributes the most, in which the effects of biased word would be more significant. For convenience, we define the examples in which the biased word makes the greatest contribution as \textbf{focus-biased examples}, and we present the statistics of biased examples and focus-biased examples in Tab.~\ref{tab:Focus-bias-0 examples statics.} and Tab.~\ref{tab:Focus-bias-1 examples statics.} (see App.~\ref{app:statistics of bias and focus-bias}). 
To measure the tendency of models' prediction,
we define $T_{c_m}$ as the tendency of model to predict of category $c_m$:
\begin{equation}
\centering 
T_{c_m} = \frac{|S_{pred}(c_m)|/|D|}{|S_{true}(c_m)|/|D|} = \frac{|S_{pred}(c_m)|}{|S_{true}(c_m)|} \qquad c_m \in (0,1)
\label{eq:influ1}
\end{equation}
where $|D|$ represents the number of observed examples, $|S_{true}(c_m)|$ and $|S_{pred}(c_m)|$ represent the number of examples with true label $c_m$ and predicted label as $c_m$ respectively. Specially, we observe the tendency of models' prediction on ``normal" biased examples and focus-biased examples, and denote them as $T_{c_m}$ and $T_{c_m}^{focus}$.
% \paragraph{Tendency to predict $c_m$.} We define 
%\paragraph{Results analysis.} 
The results are shown in Fig.~\ref{fig:influence_of_bias}. Fig.~\ref{fig:lcqmc_0} to Fig.~\ref{fig:oppo_0} show the influence of biased word$_0$ on three test sets. On DuQM (Fig.\ref{fig:duqm_0}), it is obvious that $T_{0}^{focus}$ is higher than $T_0$ by averaged 7\% with all three models, which implies that when biased word$_0$ contributes the most, models have a high tendency to predict 0. The same result is shown on LCQMC$_{test}$ (Fig. \ref{fig:lcqmc_0}). However, on OPPO (Fig.\ref{fig:oppo_0}), $T_0$ is slightly higher (0.02\textasciitilde0.05) than $T_0^{focus}$. We suppose that it is affected by the co-influencing of another shortcut and we provide an extensive experiment to discuss it in Sec.~\ref{sec:word-overlap bias}. Fig.~\ref{fig:lcqmc_1} to Fig.~\ref{fig:oppo_1} show the influence of biased word$_1$. As shown in Fig.~\ref{fig:oppo_1}, models tend to predict 1 when they concentrate on biased word$_1$ on OPPO, that $T_{1}^{focus}$ is higher than $T_{1}$ by averaged 26\% with all three models. The comparison results on DuQM(Fig.\ref{fig:duqm_1}) show the same tendency for all three models, that $T_{1}^{focus}$ is higher than $T_{1}$ by averaged 6\%. On LCQMC$_{test}$ (Fig.\ref{fig:lcqmc_1}), $T_{1}^{focus}$ is almost close to $T_{1}$ with all three models. 

Overall, we observe that when models pay more attention to biased words, they tend to assign labels over-relying on the biased words. Moreover, to explore why the tendency to 0 is not obvious on OPPO (Fig. \ref{fig:oppo_0}), we provide a further discussion about the influence of another shortcut \textit{word-overlap}.

% although not significantly
% And biased words always have a more significant impact on the out-of-domain test set (DuQM and OPPO), while the impact on the in-domain test set(LCQMC$_{test}$) seems to be slight.
 
% In real-world scenarios, the mechanism of a model's decision is complicated. Different shortcuts may interact together to give the final prediction. Here we argue that QM models are also affected by \textbf{word overlap} shortcut.

\subsection{Word-Overlap: Another Shortcut for QM Models}\label{sec:word-overlap bias}
 In real-world scenarios, different shortcuts may interact together to affect the final prediction. Word overlap  shortcut has been widely discussed in many MRC and NLI works~\cite{mccoy2019right,lai2021machine,kaushik2018much}. For QM task, the models tend to predict 0 if a sentence pair has low word overlap, i.e., there are few common words between them, and vice versa. As the result of OPPO shown in Tab.~\ref{tab:oppo_result}, even if models focus on biased word$_0$, the tendency to $0$ is not significant. We attribute the phenomenon to the word-overlap shortcut in the QM task. To eliminate the influence of word-overlap, we design an experiment on the examples in which the question pairs with high word-overlap. We use $Levenshtein Edit$ distance to measure the overlapping degree. 
%The long edit distance examples with category $0$ suffers from word-overlap shortcut.
We report the models' prediction tendency with short edit distance in Tab.~\ref{tab:oppo_result}. The results reflect that models have a higher tendency to predict $0$ on focus-biased examples than ``normal" biased examples, which implies that models tend to predict 0 if we try to eliminate the word-overlap shortcut. Specifically, compared with ``normal" biased examples, the average $T_{0}^{focus}$ of three models with edit distance $1$ increases by 0.086, which is 0.040, 0.028, 0.027 and 0.022 for edit distance of 2, 3, 4 and 5. 

% The less word-overlap the sample has, the more significant the impact of feature-label correlation is.

Generally, we can deduce that models tend to assign labels relying on the feature-label correlation trick. By eliminating the influence of word-overlap, the models' prediction tendency towards $0$ becomes significant on OPPO. Besides the spurious correlations we study in this work, NLP models are also affected by many other shortcuts.

\section{\textit{Less-Learn-Shortcut}: A Training Strategy to Mitigate Models’ Over-Reliance on Feature-Label Correlation}\label{how_to_migiate}

% 1. 方案的通用性
% 2. 在不同模型/任务/数据集/语言上都有效 broad applicability; generalizable across different xxxx
% 3. LLS decreases reliance on a well-known, lexical-overlap-driven inference heuristic for NLI.
%To mitigate models' reliance on bias, some existing work adding (task/application/model-specific) adversarial examples into the training set at the time of training. 

% In Sec.~\ref{chapter: influenece of bias} we observe that models tend to assign labels over-relying on the spurious feature-label correlation. 

In Sec.~\ref{chapter: influenece of bias} we observe that the spurious feature-label correlation will affect models' learning and deciding. To mitigate the models' shortcut learning behavior, we propose a training strategy \textit{Less-Learn-Shortcut (LLS)}, with which all the biased training examples are penalized according to their biased degrees (in Sec.~\ref{reweight}) during fine-tuning. Most of the existing strategies to mitigate shortcut learning include data augmentation~\cite{jin2020bert,alzantot2018generating} and adversarial training~\cite{stacey2020avoiding}, which are task-relevant. Our proposed method \textit{LLS} is task-agnostic and can be easily transferred to different NLP tasks.

% Therefore, we compare our $LLS$ method with some task-agnostic methods (in Sec.~\ref{experiment}).
%tend to overfit the superficial cues in the training data and make prediction 

% and achieve improvements on the constructed adversarial test set
% Most of previous works mitigate shortcut by using their analytical phenomenon to construct adversarial data only for specific dataset or task. Obviously, this line of work does not migrate well to other datasets or tasks and requires extra manpower or cost to construct adversarial data. In our work, according to our above analysis on QM task, we score examples and penalize losses with their scores as weights for mitigating the model's reliance on bias. We verify the effectiveness and generalization of our method on QM and NLI task. Besides, our method doesn't need extra manpower or cost. Here we show the process of scoring and the results of our experiment. 

\subsection{Reweight Biased Examples}\label{reweight}

% spurious correlations between shortcut features and labels

To mitigate the models' over-reliance on the feature-label correlations, a straightforward idea is to down-weight the biased examples, so that the models are prevented from over-fitting the spurious correlations. In this section, we will introduce how we reweight the biased examples.

% and forced to learn semantic features
\paragraph{Quantify the impact of correlation.}
In Sec.~\ref{set_up}, we have defined \textit{biased degree} $d_{w_i}^{c_m}$ to measure the correlation between the word $w_i$ and the label $c_m$, which can quantify the impact of the correlation. The maximum biased degree of a word among all categories is denoted as $b_{w_i}^{\circ}$ (C represents all categories).

\begin{equation}\label{eqt:LLS_only_bias}
b_{w_i}^{\circ}=\max d_{w_i}^{c_m}, c_m \in C
\end{equation}

Furthermore, some existing works show that the word frequency in the training data also influences the models' prediction~\cite{gu2020token,cui2016attention,ott2018analyzing}. Considering both biased degree and word frequency, we formulate the impact of a biased word as

\begin{equation}\label{eqt:LLS_b_f}
b_{w_i}=\max d_{w_i}^{c_m}+\alpha f_{w_i}, c_m \in C
\end{equation}
where $f_{w_i}$ represents the frequency of words $w_i$ occurring in the training dataset, and $\alpha$ is a trade-off factor. Then the impact of a biased example can be formulated as the average impact of all biased words it contains:
% biased words are words highly correlated with a specific label, and biased examples are examples containing at least one biased word. 

%To quantify the impact of bias, we quantify the impact as 

\begin{equation}\label{eqt:LLS_exp}
b_{e}^{\circ}=\frac{1}{n}\sum\limits_{i=1}^n b_{w_i}
\end{equation}

\begin{table*}[ht]
\begin{spacing}{1.1}
\centering
\small
{
\begin{tabular}{ll|ccc|cc|cc}
\toprule[0.7pt]
\multicolumn{2}{c}{\textbf{Task}}            & \multicolumn{3}{c}{\textbf{QM}}      & \multicolumn{2}{c}{\textbf{NLI}}                             & \multicolumn{2}{c}{\textbf{SA}} \\
\midrule[0.5pt]
\textbf{Model}                    &    \textbf{Strategy}      & \textbf{LCQMC$_{test}$}   & \textbf{DuQM}    & \textbf{OPPO}    & \textbf{SNLI$_{test}$} &\textbf{HANS$_{test}$} & \textbf{Chn$_{test}$}        & \textbf{SENTI$_{robust}$}     \\
\midrule[0.5pt]
\multirow{6}{*}{BERT}    & \textit{Finetune} & 87.16\% & 67.99\% & 81.99\% & 90.80\%                  & 57.71\%                  & \textbf{95.53\%}    & 65.77\%   \\
& {\textit{Rew.$_{bias}$}}      & 87.20\% & 67.79\% & 81.80\% & 90.50\%                  & 59.12\%                  & 94.89\%    & 67.28\%   \\
& {\textit{Forg.}}      & 86.84\% & 68.20\% & 81.57\% & 90.48\%                  & \textbf{61.51\%}                & 94.92\%    & 66.88\%   \\
\cline{2-9}
                         & {\textit{LLS$_d$}}      & 87.40\% & 68.52\% & 81.91\% & 90.69\%                  & 59.10\%                  & 95.50\%      & \textbf{67.51\%}     \\
                         & {\textit{LLS$_{d+f}$}}      & 87.27\% & 69.05\% & \textbf{82.08\%}  & 90.57\%                  & 59.58\%                  & 95.00\%    & 67.44\%   \\
                         & {\textit{LLS}}      & \textbf{87.86\%}  & \textbf{69.20\%}  & 81.84\% & \textbf{90.82\%}                   & 59.33\%                  & -           & -          \\
\midrule[0.5pt]
\multirow{6}{*}{ERNIE}   & \textit{Finetune} & 87.63\% & 70.08\% & 82.56\% & 91.19\%                  & 62.59\%                  & \textbf{96.08\%}    & 63.45\%   \\
& {\textit{Rew.$_{bias}$}}      & 87.09\% & \textbf{71.68\%} & 82.44\% & 91.28\%            & 62.88\%                  & 95.64\%    & 63.80\%   \\
& {\textit{Forg.}}      & 87.04\% & 71.62\% & 82.50\% & 90.91\%                  & \textbf{65.22\%}           & 95.11\%    & 63.52\%   \\
\cline{2-9}
                         & {\textit{LLS$_d$}}      & 87.38\% & 70.61\% & 82.30\% & 91.27\%                  & 64.26\%                  & 95.83\%    & \textbf{64.18\%}    \\
                         & {\textit{LLS$_{d+f}$}}      & 87.61\% & 70.88\% & \textbf{82.61\%}  & \textbf{91.31\%}                   & 64.55\%                  & 96.00\%     & 63.55\%   \\
                         & {\textit{LLS}}      & \textbf{88.16\%}  & 71.65\%  & 82.52\% & 91.12\%                  &  64.76\%                   & -           & -          \\
\midrule[0.5pt]
\multirow{6}{*}{RoBERTa} & \textit{Finetune} & 87.58\% & 72.86\% & 82.60\% & \textbf{92.54\%}     & 74.31\%                  & 95.00\%    & 65.32\%   \\
& {\textit{Rew.$_{bias}$}}      & 87.68\% & 73.91\% & 82.70\% & 92.41\%            & 73.14\%                  & 95.22\%    & 65.92\%   \\
& {\textit{Forg.}}      & 86.50\% & 73.76\% & 82.50\% & 92.25\%                  & 74.65\%                  & 94.33\%    & 66.07\%   \\
\cline{2-9}
                         & {\textit{LLS$_d$}}      & 87.85\% & 74.14\% & 82.80\% & 92.47\%                   & 74.52\%                  & \textbf{95.64\%}     & 66.17\%   \\
                         & {\textit{LLS$_{d+f}$}}      & 87.84\% & 73.42\% & 82.71\% & 92.44\%                  & 73.74\%                  & 95.28\%    & \textbf{66.46\%}    \\
                         & {\textit{LLS}}      & \textbf{88.46\%}  & \textbf{74.18\%}  & \textbf{82.81\%}  & 92.42\%                  & \textbf{74.88\%}                   & -           &  -         \\
\bottomrule[0.7pt]
\end{tabular}
}
\caption{Performance (accuracy\%) of three models on LCQMC$_{train}$, SNLI$_{train}$ and Chnsenticorp$_{train}$ respectively. As SA task is a single-sentence classification task which not struggle with word-overlap shortcut, we only try \textit{LLS$_d$} and \textit{LLS$_{d+f}$} strategy on it. We select \textit{Finetune}, \textit{Rew.$_{bias}$} and \textit{Forg.} strategies as our baselines. For each model, bold font represents the best performance.}
\label{tab:perfor_comparision}
\end{spacing}
\end{table*}

\paragraph{Exclude the impact of word-overlap for sentence pair tasks.}\label{word_overlap}

% In Sec.~\ref{sec:word-overlap bias}, we observe that word-overlap is another shortcut in the sentence pair tasks: models tend to give positive predictions to sentence pairs with high word-overlap and negative to low word-overlap. To exclude the impact of word-overlap, when two shortcuts conflict in one biased example\footnote{Two shortcuts conflict when the biased word${_1}$ (tend to predict label 1) only occurs in one sentence (tend to predict label 0), or the biased word${_0}$ (tend to predict label 0) occurs in both sentences (tend to predict label 1).}, we only consider the impact of the biased word with minimum biased degree:

In Sec.~\ref{sec:word-overlap bias}, we observe that word-overlap is another shortcut in the sentence pair tasks: models tend to give positive predictions to sentence pairs with high word-overlap and negative to low word-overlap. To exclude the impact of word-overlap, when two shortcuts conflict in one biased example, we only consider the impact of the biased word with minimum biased degree:

\begin{equation}\label{eqt:LLS_exp_2}
 b_{e}=\begin{cases} \min b_{w} &, \ conflict\\\frac{1}{n}\sum\limits_{i=1}^n b_{w_i} & , otherwise\end{cases} 
\end{equation}

% in [0,1]
\paragraph{Calculate the loss weight of the biased examples.}
We rescale $b_e$ of all biased examples with the min-max normalization. The loss weights of the biased examples can be denoted as follows, with which the biased examples with higher $b_e$ will be assigned smaller loss weights:

%For biased examples, we obtain the score set Escore=\{Escore$_{e_0}$,...,Escore$_{e_m}$\}. We then normalize Escore$_{e_i}$ as weight to penalize loss and the denote is as
\begin{equation}\label{eq:weight}
w_{e_j}=1-\beta\frac{b_{e_j}-\min b_{e}}{\max b_{e}-\min b_{e}}
\end{equation}
where $e_j$ belongs to biased example and $\beta$ is used for adjusting the low bound of the normalized interval, and the high bound is fixed to 1. It is worth noting that we only reweight the biased examples, and the loss weights of unbiased examples are 1.
%where $\beta$ is used for adjusting the low bound of the normalized interval, the high bound is fixed at 1.

%Now, we get the weight w$_{e_i}$ for each example. We weight original loss for each example to calculate our loss and the denote is as
%\begin{equation}
%loss_{our}=w_{e_i}\times loss_{original}
%\end{equation}

\subsection{Experimental Results}\label{experiment}

% Chn$_{test}$ represents Chnsenticorp$_{test}$.
% We first introduce our compared baseline, then conduct experiments on QM task, NLI task (sentence pair classification task) and SA task (single sentence classification task). All experimental settings are introduced in App.\ref{exp_setting}. We report average experimental results with three different seeds and our performance improvements are statistically significant with p-value of paired t-test less than 0.05.

First, we introduce our comparison baseline and then conduct experiments on QM task, NLI task (sentence pair classification task), and SA task (single sentence classification task). Detailed information about the experimental settings can be found in App.\ref{exp_setting}. We present the average results of three different seeds and our performance improvements are statistically significant with a p-value of paired t-test less than 0.05.

% in the majority of all cases

% in Tab.\ref{tab:perfor_comparision}

\begin{table}[t]
\begin{spacing}{1.1}
\centering
\small
{
\begin{tabular}{l|c|c|c}
\toprule[0.7pt]
{\textbf{Dataset}} & {\textbf{\#\ Words}} & {\textbf{\#\ B-word}} & {\textbf{\%B-word}}\\
\midrule[0.7pt]
{SNLI$_{train}$} & {42,567} & {1,261} & {2.96\%} \\
{MNLI$_{train}$} & {101,705} & {202} & {0.22\%} \\
{Chnsenticorp$_{train}$} & {35,274} & {4,956} & {14.05\%}\\
\bottomrule[0.7pt]
\end{tabular}
}
\caption{The statistics of biased words in SNLI$_{train}$, MNLI$_{train}$ and Chnsenticorp$_{train}$. B-word represents biased word.}
\label{tab:nli_bias_word_statics}
\end{spacing}
\end{table}

% Question Matching (QM), Natural Language Inference (NLI) (sentence pair tasks), and Sentiment Analysis (SA) (single sentence task).
\paragraph{Baseline.} 
In addition to select \textit{Finetune} as our baseline, we re-implement \textit{Rew.$_{bias}$} ~\cite{utama2020towards,clark2019don} and \textit{Forg.}~\cite{yaghoobzadeh2021increasing} strategies. The core idea of \textit{Rew.$_{bias}$} is similar to our \textit{LLS} strategy, which is down-weighting biased examples. \textit{Rew.$_{bias}$} needs to additionally train a bias-only model to score biased examples. \textit{Forg.} strategy uses examples forgotten by the model during training to do secondary training. The above two strategies are both task-agnostic and can be applied to any NLP task.

% , therefore, we compare our $LLS$ method with them.

\paragraph{QM task.} 
As shown in Tab.~\ref{tab:perfor_comparision}, for BERT and RoBERTa, our \textit{LLS} strategy performs best on both the in-domain LCQMC$_{test}$ (87.86\% and 88.46\%) and adversarial DuQM (69.20\% and 74.18\%). For ERNIE, our \textit{LLS} strategy performs best on in-domain LCQMC$_{test}$ (88.16\%) and performs close to best on the adversarial DuQM (71.65\%). Furthermore, we observe that although \textit{Rew.$_{bias}$} and \textit{Forg.} strategies improve the model performance on the adversarial DuQM, they only remain performance on LCQMC$_{test}$ and OPPO. By contrast, our \textit{LLS} strategy can improve the model performance on all three test sets.

% our $LLS$ method significantly improves models' performance on the in-domain LCQMC$_{test}$ set while improving their performance better on the adversarial dataset DuQM.Compared to finetune, the performance of models with our $LLS$ method have a significant improvement.

% Rew.$_{bias}$ and Forg. methods improve models' performance on the adversarial dataset DuQM and almost remain their performance on the in-domain LCQMC$_{test}$ set and out-of-doamin OPPO set. Compared to them, our $LLS$ method significantly improves models' performance on the in-domain LCQMC$_{test}$ set while improving their performance better on the adversarial dataset DuQM. Specifically, for BERT and RoBERTa, our LLS method performs best on both the in-domain LCQMC$_{test}$ (87.86\% and 88.46\%) and adversarial dataset DuQM (69.20\% and 74.18\%). For ERNIE, our LLS method performs best on the in-domain LCQMC$_{test}$ (88.16\%) and perform close to best on adversarial dataset DuQM (71.65\%).

\begin{table}[t]
\begin{spacing}{1.1}
\centering
\small
{
\begin{tabular}{l|c|c|c}
\toprule[0.7pt]
{\textbf{Dataset}} & {\textbf{\#\ Examples}} & {\textbf{\#\ B-exp}} & {\textbf{\%B-exp}}\\
\midrule[0.7pt]
{SNLI$_{train}$} & {549,367} & {26,590} & {4.84\%} \\
{MNLI$_{train}$} & {392,702} & {993} & {0.25\%} \\
{Chnsenticorp$_{train}$} & {9,600} & {9,151} & {95.11\%}\\
\bottomrule[0.7pt]
\end{tabular}
}
\caption{The statistics of biased examples in SNLI$_{train}$, MNLI$_{train}$ and Chnsenticorp$_{train}$. B-exp represents biased example.}
\label{tab:nli_bias_example_statics}
\end{spacing}
\end{table}

To better investigate the contributions of different components of \textit{LLS}, we compare \textit{LLS} with two ablations: \textit{LLS$_d$} only employs the biased degree to measure the impact of correlation, and does not consider the impact of word-overlap (Eq.~\ref{eqt:LLS_only_bias}, \ref{eqt:LLS_exp}, and \ref{eq:weight}); \textit{LLS$_{d+f}$} considers both biased degree and word frequency to measure the correlation, but also does not consider the impact of word-overlap (Eq.~\ref{eqt:LLS_b_f}, \ref{eqt:LLS_exp}, and \ref{eq:weight}). As shown in Tab.~\ref{tab:perfor_comparision}, \textit{LLS} generally performs the best, indicating that considering word frequencies and excluding word-overlap has a positive effect.

% performs best generally, which indicates a positive effect with consideration of word frequencies and exclusion of word-overlap. 

\paragraph{NLI task.}\label{nli_task}
NLI task aims to determine the relationship between two sentences, whether a premise sentence entails a hypothesis sentence. It is normally formulated as a multi-class classification problem. In our experiments, we try two NLI datasets as the training sets, SNLI~\cite{bowman2015large} and MNLI~\cite{williams2017broad}. Tab.~\ref{tab:nli_bias_word_statics} and~\ref{tab:nli_bias_example_statics} give the statistics of SNLI$_{train}$ and MNLI$_{train}$. Although only 2.96\% of words in SNLI$_{train}$ are biased words, they occur in 4.84\% of examples. Compared to SNLI$_{train}$, MNLI$_{train}$ is relatively unbiased and contains only 202 biased words (0.22\%) and 993 biased examples (0.25\%). 

We first conduct our experiment on SNLI$_{train}$. We train models on SNLI$_{train}$ and evaluate them on the in-domain SNLI$_{test}$ and the adversarial HANS$_{test}$. SNLI is a dataset with three classes: entailment, neutral, and contradiction. HANS is a two-class dataset, entailment and non-entailment. As done in previous work~\cite{mccoy2019right}, to evaluate models on HANS$_{test}$, we convert neutral or contradiction labels to non-entailment. The experimental results are shown in Tab.~\ref{tab:perfor_comparision}. For BERT and ERNIE, \textit{Forg.} strategy improves the model performance more significantly on the adversarial HANS$_{test}$. We present the statistics of forgotten examples (see App.~\ref{forg_sample}), and observe that for the large-scale SNLI$_{train}$, small models such as BERT and ERNIE are more likely to forget examples. Therefore, secondary training with forgotten examples can better help small models increase their robustness. In contrast, for the large RoBERTa model, \textit{Forg.} strategy yields little and our \textit{LLS} strategy performs better. Furthermore, compared to \textit{Finetune} and \textit{Rew.$_{bias}$} strategies, for all three models, our \textit{LLS} strategy obtains a more significant benefit on the adversarial HANS$_{test}$ while maintaining good performance on the in-domain SNLI$_{test}$.

% For RoBERTa, our \textit{LLS} strategy performs better than \textit{Forg.} strategy. Meanwhile, compared to \textit{Finetune} and \textit{Rew.$_{bias}$} strategies, for all three models, our \textit{LLS} strategy obtains a more significant benefit on the adversarial HANS$_{test}$.

% As shown in App.~\ref{forg_sample}, we present the statistics of forgotten examples with \textit{Forg.} strategy. We observe that for the large-scale SNLI$_{train}$, small models such as BERT and ERNIE are more prone to forgetting examples. Therefore, secondary-training with forgotten examples can better help small models increase their robustness. By contrast, for RoBERTa, our \textit{LLS} strategy performs better than \textit{Forg.} strategy. Meanwhile, compared to \textit{Finetune} and \textit{Rew.$_{bias}$} strategies, for all three models, our \textit{LLS} strategy obtains a more significant benefit on the adversarial HANS$_{test}$. 

% However, models' performance on the in-domain SNLI$_{test}$ set only remains, and even decreases slightly. 

The results on MNLI$_{train}$ are shown in Tab.~\ref{tab:mnli_performance_comparision}. Due to the fact that the MNLI$_{train}$ contains fewer biased examples, the effect of \textit{LLS} is not significant. This suggests that \textit{LLS} strategy is more effective for the training set with biased data distribution, helping models learn the spurious correlation less.

% Compared to the results of SNLI$_{train}$, the effect of \textit{LLS} is not significant on MNLI$_{train}$. We believe this is due to the fact that the MNLI$_{train}$ contains fewer biased examples. Therefore, our strategy is more suitable for the training set with the biased data distribution, where \textit{LLS} can help models learn the spurious correlation less.

% We further validate our guess in Sec.~\ref{sa_task}.

\begin{table}[ht]
\begin{spacing}{1.1}
\centering
% \vspace{-0.cm}
\small
{
\begin{tabular}{llcc}
\toprule[0.7pt]
{\textbf{Model}}& {\textbf{Strategy}} & {\textbf{MNLI$_{test}$}} & \textbf{HANS$_{test}$} \\
\midrule[0.7pt]
\multirow{4}{*}{BERT} &  {\textit{Finetune}} & {84.22} & {52.01} \\
{} &  {\textit{LLS$_d$}} & {84.36} & \textbf{52.40} \\
{} &  {\textit{LLS$_{d+f}$}}& {84.31} & {51.99} \\
{} &  {\textit{LLS}}& \textbf{84.49} & 52.24 \\
\bottomrule[0.7pt]
\end{tabular}
}
\caption{Performance (accuracy\%) of BERT on MNLI$_{train}$.}
\label{tab:mnli_performance_comparision}
\end{spacing}
\end{table}

\paragraph{SA task.}\label{sa_task}
SA task aims to determine whether a sentence has a positive or negative sentiment. In our experiment, we train models on Chnsenticorp$_{train}$\footnote{https://github.com/pengming617/bert\_classification}
 and evaluate them on the in-domain Chnsenticorp$_{test}$ and the adversarial SENTI$_{robust}$~\cite{wang2021dutrust}. As shown in Tab.~\ref{tab:nli_bias_word_statics} and Tab.~\ref{tab:nli_bias_example_statics}, 14.05\% of words in Chnsenticorp$_{train}$ are biased words, and they appear in 95.11\% of the examples. Unlike NLI and QM tasks, SA task is a single-sentence classification task that is not affected by the word-overlap shortcut, thus we only report the results of \textit{LLS$_{d}$} and \textit{LLS$_{d+f}$} in Tab.~\ref{tab:perfor_comparision}. Compared to \textit{Rew.$_{bias}$} and \textit{Forg.} strategies, our \textit{LLS$_{d}$} and \textit{LLS$_{d+f}$} strategies obtain a more significant benefit on the adversarial SENTI$_{robust}$ and perform better on the in-domain Chnsenticorp$_{test}$. Additionally, it is worth noting that for the small-scale Chnsenticorp${train}$, models will not forget too many samples (see App.~\ref{forg_sample}) and \textit{Forg.} strategy yields little.

In summary, our proposed \textit{LLS} strategy can significantly improve the model performance on adversarial data while maintaining good performance on in-domain data. Our experiments show that existing strategies struggle to stably improve performance on in-domain data, making further research necessary. Furthermore, we reveal scenarios in which these strategies are applicable. Compared to the \textit{Rew.$_{bias}$} strategy, \textit{LLS} strategy demonstrates greater advantages on various tasks. However, \textit{LLS} strategy is not applicable for the relatively unbiased dataset, such as MNLI$_{train}$. On the other hand, \textit{Forg.} strategy shows its own advantages on SNLI$_{train}$. Specifically, when training a small model on a large-scale dataset, \textit{Forg.} strategy is a good option to consider.

\section{Conclusion}\label{conclusion}
In this paper, we explore models' shortcut learning behavior of spurious correlations between features and labels, and propose a training strategy \textit{LLS} to mitigate the over-reliance of NLP models on the shortcut. Specifically, we observe that the models are prone to learn spurious correlations, and the biased words make significantly higher contributions to models' predictions than random words. Moreover, we observe that the models tend to be misled by biased words to assign labels. To mitigate the over-reliance on biases, we propose a training strategy \textit{LLS} to penalize the shortcut learning behavior of models. Experimental results show that \textit{LLS} can improve the model performance on adversarial data while keeping good performance on in-domain data, and it is task-agnostic, which can be easily transferred to other tasks. In future research, we will explore how to better measure and formalize the shortcuts in the training data and generalize them as a class of problems.

\appendix
% \clearpage

\section{Data Statistics}\label{app:lcqmc statics}
Data statistics are presented in Tab.~\ref{tab:dataset_basic_statics}.

\begin{table}[h]
\begin{spacing}{1.1}
\centering
% \scalebox{0.65}
\setlength{\tabcolsep}{1mm}{
\small{
\begin{tabular}{lccc|ccc}
\toprule[0.7pt]
\multirow{2}{*}{\textbf{Dataset}} & \multicolumn{2}{c}{\textbf{Word cnt.}} & \multirow{2}{*}{\textbf{Total}} & \multicolumn{2}{c}{\textbf{Category}} & \multirow{2}{*}{\textbf{Total}} \\
{} & q1 & q2 & {} & {\#0} & {\#1} & {} \\
\midrule[0.7pt]
{L$_{train}$} & 6.04 & 6.36 & 12.40 & 100,192 & 138,574 & 238,766\\
{L$_{test}$} & 5.51 & 5.61 & 11.12 & 6,250 & 6,250 & 12,500\\
{DuQM} & 4.66 & 4.80 & 9.46 & 7,318 & 2,803 & 10,121\\
{OPPO} & 4.82 & 4.71 & 9.53 & 7.160 & 2.840 & 10,000\\
\bottomrule[0.7pt]
\end{tabular}
}}
\caption{Data statistics. L$_{train}$ denotes LCQMC training set, and L$_{test}$ denotes LCQMC test set.}
\label{tab:dataset_basic_statics}
\end{spacing}
\end{table}

\section{Training Loss}\label{app:training loss}
\begin{figure}[h]
\centering
    \subfigure[Training loss curve of BERT.]{
    \centering
    \includegraphics[scale=0.25]{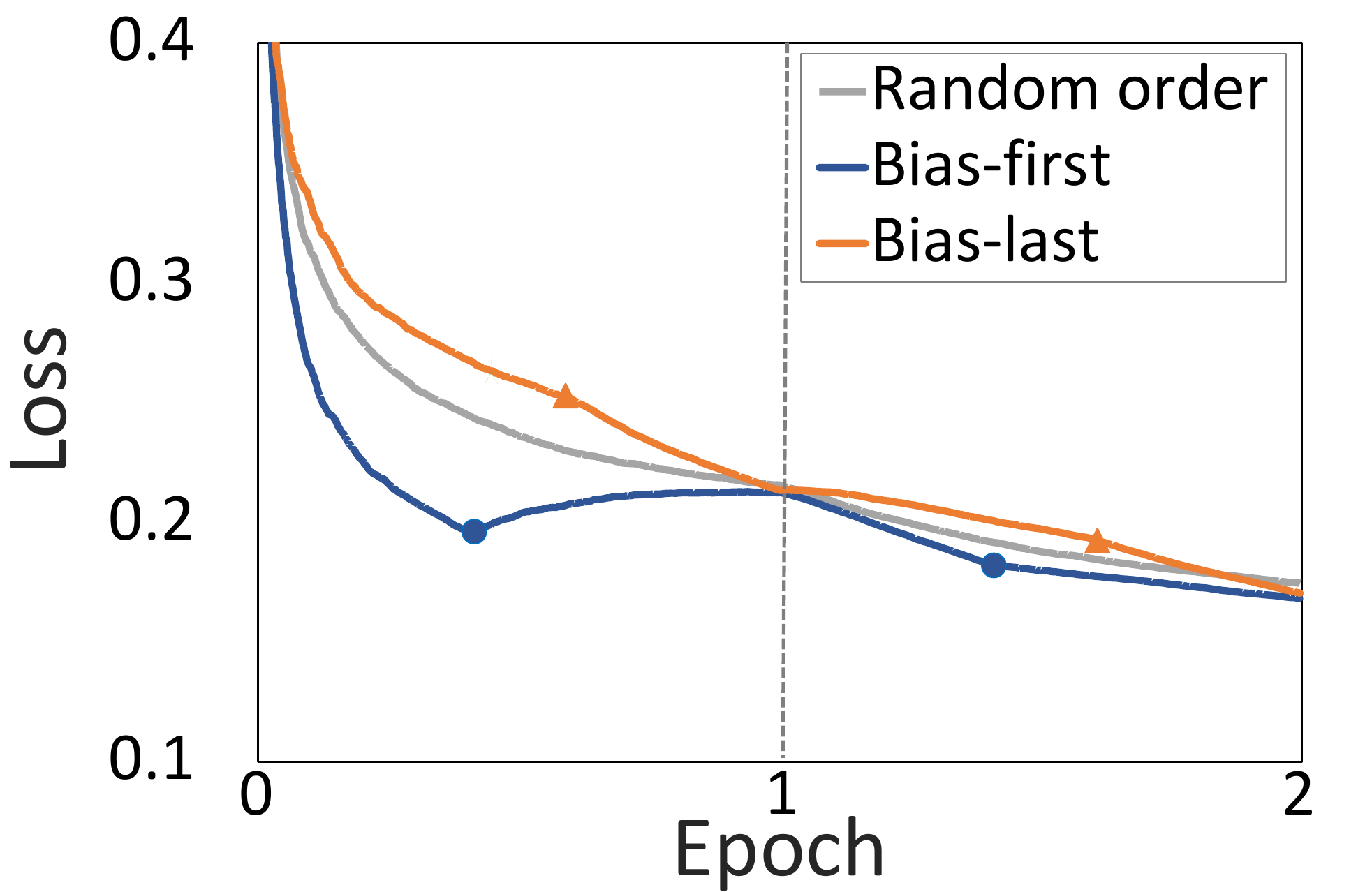}
}
\centering
    \subfigure[Training loss curve of ERNIE.]{
    \centering
    \includegraphics[scale=0.25]{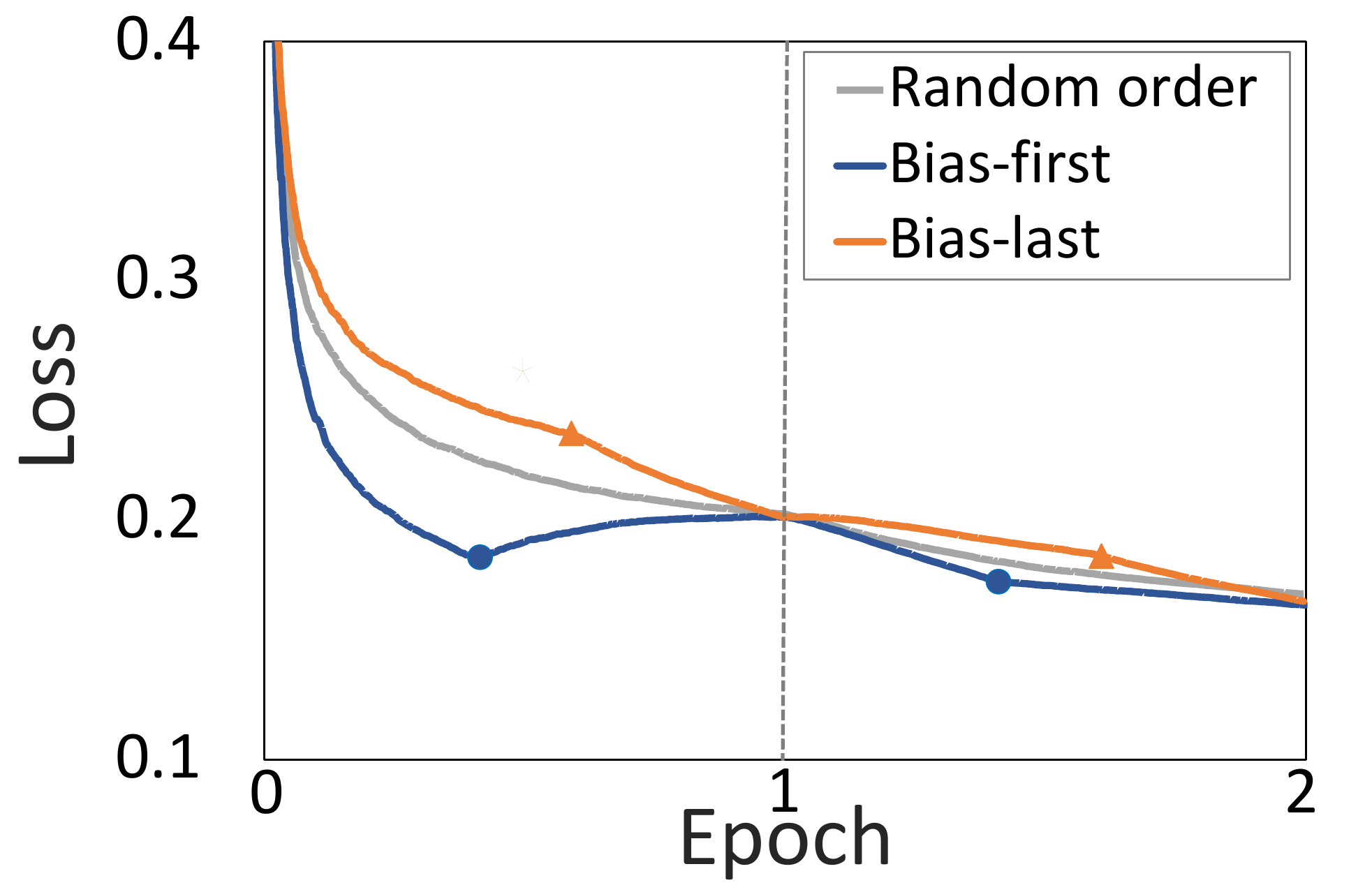}
}
\caption{Training loss curves of BERT and ERNIE on LCQMC$_{train}$, in which \textcolor{blue}{$\bullet$} represents finishing learning biased examples, and \textcolor{orange}{$\blacktriangle$} represents finishing learning unbiased examples.}
\label{fig:training loss curve app}
\end{figure}

\section{Examples of Bias-word}\label{app:example of bias-word} 
Examples of bias-word$_0$ and bias-word$_1$ are given in Tab.~\ref{tab:bias_word_illustraion}.

\begin{table}[h]
% \begin{spacing}{1.1}
\centering
\newcommand{\tabincell}[2]{\begin{tabular}{@{}#1@{}}#2\end{tabular}}
\small
{
\begin{tabular}{c|c|cccc}
\toprule[0.7pt]
{} & \multirow{2}{*}{\textbf{Word}} & \multicolumn{2}{c}{\textbf{Category}} & \multirow{2}{*}{\textbf{Total}} & \multirow{2}{*}{\textbf{B-degree}} \\
{} & {} & {\#0} & {\#1} & {} & {}\\
\midrule[0.7pt]

\multirow{2}{*}{\textbf{B-word$_{0}$}} &     \tabincell{c}{\begin{CJK*}{UTF8}{gbsn}
           漂浮
        \end{CJK*} \\ (float)} & {5} & {0} & {5} & {1.00} \\
\midrule[0.5pt]
\multirow{2}{*}{\textbf{B-word$_{1}$}} & \tabincell{c}{简便 \\ (handy)} & {2} & {33} & {35} & {0.94}\\
\bottomrule[0.7pt]
\end{tabular}
}
\caption{Examples of bias-word$_0$, and bias-word$_1$. B-word$_{0}$ represents bias-word$_0$, and B-word$_{1}$ represent bias-word$_1$.}
\label{tab:bias_word_illustraion}
% \end{spacing}
\end{table}

\section{Statistics of Bias-Example and Focus-Bias Examples}\label{app:statistics of bias and focus-bias}
The statistics of "normal" bias-examples and focus-bias examples are given in Tab.~\ref{tab:Focus-bias-0 examples statics.} and Tab.~\ref{tab:Focus-bias-1 examples statics.}.

\begin{table}[h]
% \begin{spacing}{1.2}
\centering
\setlength{\tabcolsep}{1.5mm}{
\scriptsize
{\begin{tabular}{lcc|cc|cc}
\toprule[0.7pt]
\multirow{2}{*}{\textbf{B-word$_0$}} & \multicolumn{2}{c|}{\textbf{LCQMC$_{test}$}} & \multicolumn{2}{c|}{\textbf{DuQM}} & \multicolumn{2}{c}{\textbf{OPPO}} \\
& \#\ S  &\#\ $S_{focus}$ &\#\ S & \#\ $S_{focus}$ &\#\ S & $S_{focus}$ \\
\midrule[0.7pt]
BERT & \multirow{3}{*}{1,777} &551  & \multirow{3}{*}{2,375} & 824 & \multirow{3}{*}{1,991} & 474 \\
ERNIE &  & 554 &  & 879 & & 494 \\ 
RoBERTa &  & 517 &  & 844 & & 457 \\
\bottomrule[0.7pt]
\end{tabular}
}}
\caption{Statistics of bias-example$_0$ ($S$) and focus-bias$_0$ examples ($S_{focus}$). Focus-bias$_0$ examples represent the examples where bias-word$_0$ makes great contribution.}
\label{tab:Focus-bias-0 examples statics.}
% \end{spacing}
\end{table}

\begin{table}[h]
\begin{spacing}{1.2}
\centering
% \scalebox{0.65}
\setlength{\tabcolsep}{1.5mm}{
\scriptsize
{
\begin{tabular}{lcc|cc|cc}
\toprule[0.7pt]
\multirow{2}{*}{\textbf{B-word$_1$}} & \multicolumn{2}{c|}{\textbf{LCQMC$_{test}$}} & \multicolumn{2}{c|}{\textbf{DuQM}} & \multicolumn{2}{c}{\textbf{OPPO}} \\
&\#\ S  &\#\ $S_{focus}$ &\#\ S &\#\ $S_{focus}$ &\#\ S & $S_{focus}$ \\
\midrule[0.5pt]
BERT & \multirow{3}{*}{1,543} & 753 & \multirow{3}{*}{1,095} & 336 & \multirow{3}{*}{602} & 159 \\
ERNIE &  & 774 &  & 340 & & 163 \\ 
RoBERTa &  & 722 &  & 309 & & 154 \\
\bottomrule[0.7pt]
\end{tabular}
}}
\caption{Statistics of bias-example$_1$ ($S$) and focus-bias$_1$ examples ($S_{focus}$). Focus-bias$_1$ examples represent the examples where bias-word$_1$ makes great contribution.}
\label{tab:Focus-bias-1 examples statics.}
\end{spacing}
\end{table}

\section{Experimental Settings}\label{exp_setting}
Experimental settings are introduced Tab.~\ref{app:exp_settings}.

\begin{table}[h]
\centering
\small
{
\begin{tabular}{clccccc}
\toprule[0.7pt]
           Dataset                   &   Model      & Epoch &  Lr & Bs & Wd                       \\
\midrule[0.5pt]
\multirow{3}{*}{LCQMC}        & BERT    & 2     & 2e-5            & 64         & 0.01                               \\
                              & ERNIE   & 2     & 2e-5            & 64         & 0.01                               \\
                              & RoBERTa & 3     & 5e-6            & 64         & 0.01                               \\
\midrule[0.5pt]
\multirow{3}{*}{SNLI}         & BERT    & 2     & 2e-5            & 64         & 0.01                               \\
                              & ERNIE   & 2     & 2e-5            & 64         & 0.01                               \\
                              & RoBERTa & 2     & 5e-6            & 64         & 0.01                               \\
\midrule[0.5pt]
\multirow{3}{*}{Chnsenticorp} & BERT    & 5     & 2e-5            & 64         & 0.01
                              \\
                              & ERNIE   & 5     & 2e-5            & 64         & 0.01         
                              \\
                              & RoBERTa & 5     & 5e-6            & 64         & 0.01         
                              \\
\bottomrule[0.7pt]
\end{tabular}
}
\caption{Introduce our experimental settings. Lr represents learning rate, Bs represents batch size and Wd represents weight decay.}
\label{app:exp_settings}
\end{table}

\section{Statistics of Forgotten Examples}\label{forg_sample}
In Tab.~\ref{app_forg}, we record the number of examples forgotten by models on different tasks.

\begin{table}[h]
\centering
\small
{
\begin{tabular}{cccc}
\toprule[0.7pt]
             & BERT  & ERNIE & RoBERTa \\
\midrule[0.5pt]
LCQMC        & 15,589 & 18,591 & 17,846   \\
SNLI         & 55,696 & 51,397 & 37,466   \\
Chnsenticorp & 1,154  & 1,161  & 1,187   \\
\bottomrule[0.7pt]
\end{tabular}
}
\caption{Statistics of forgotten examples during finetuning. Each value represents how many samples models forget during finetuning. An obvious phenomenon is that on large-scale SNLI dataset, small models tend to forget more samples (55,696 for BERT, 51,397 for ERNIE, but 37,466 for RoBERTa).}
\label{app_forg}
\end{table}

\clearpage

\section*{Acknowledgements}

This work was supported by the National Key R\&D Program of China [2021ZD0113302]; and the National Natural Science Foundation of China [62206079]; and Heilongjiang  Provincial Natural Science Foundation of China [YQ2022F006]. 

\bibliographystyle{named}
\bibliography{custom}
% \bibliography{ijcai23}

\end{CJK*}

\end{document}